%% file: 00_main.tex
\documentclass[11pt]{article}
\usepackage{hyperref}

\PassOptionsToPackage{hyphens}{url}
\usepackage{url}
\usepackage{naaclhlt2018}
\usepackage{times}
\usepackage[english]{babel}
\usepackage{fontenc}
\usepackage[utf8x]{inputenc}
\usepackage{enumitem}

\usepackage{amsmath}
\usepackage{graphicx}
\usepackage{booktabs}

\aclfinalcopy

\newcommand{\tweetquote}[1]{{\color{teal}\textit{``#1''}}}

\newcommand{\nU}{\ensuremath n^{(u)}}
\newcommand{\NU}{\ensuremath N^{(u)}}
\newcommand{\phu}{\ensuremath \hat{p}^{(u)}}
\newcommand{\nca}{\ensuremath \nU_{\textsc{ca}}}
\newcommand{\nes}{\ensuremath \nU_{\textsc{es}}}
\newcommand{\pro}[1]{\ensuremath #1_{\textit{pro}}}
\newcommand{\anti}[1]{\ensuremath #1_{\textit{anti}}}
\newcommand{\npro}{\ensuremath \pro{\NU}}
\newcommand{\nanti}{\ensuremath \anti{\NU}}

\title{Sí o no, ¿què penses? \\Catalonian Independence and Linguistic Identity on Social Media}

\author{Ian Stewart\thanks{\enspace Equal contributions.}\: and Yuval Pinter\footnotemark[1]\: and Jacob Eisenstein \\
  School of Interactive Computing \\
  Georgia Institute of Technology \\
  Atlanta, GA, USA \\
  \{istewart6, uvp, jacobe\}@gatech.edu}

\begin{document}

\maketitle

\begin{abstract}
Political identity is often manifested in language variation, but the relationship between the two is still relatively unexplored from a quantitative perspective.
This study examines the use of Catalan, a language local to the semi-autonomous region of Catalonia in Spain, on Twitter in discourse related to the 2017 independence referendum.
We corroborate prior findings that pro-independence tweets are more likely to include the local language than anti-independence tweets.
We also find that Catalan is used more often in referendum-related discourse than in other contexts, contrary to prior findings on language variation.
This suggests a strong role for the Catalan language in the expression of Catalonian political identity.
\end{abstract}

\input{01_intro}

\input{02_setup}

\input{03_exp1}

\input{04_exp2}

\input{05_conclusion}

\input{06_acknowledgments}

\bibliographystyle{acl_natbib.bst}
\bibliography{main}

\end{document}

%% file: 01_intro.tex
\section{Introduction}

\input{t01_hashtag_tables}

Social identity is often constructed through language use, and variation in language therefore reflects social differences within the population~\cite{labov1963}.
In a multilingual setting, an individual's preference to use a local language rather than the national one may reflect their political stance, as the local language can have strong ties to cultural and political identity~\cite{moreno1998,crameri2017}.
The role of linguistic identity is enhanced in extreme situations such as referenda, where the voting decision may be driven by identification with a local culture or language~\cite{schmid2001}.

In October 2017, the semi-autonomous region of Catalonia held a referendum on independence from Spain, where 92\% of respondents voted for independence~\cite{fotheringham2017}.
To determine the role of the local language Catalan in this setting, we apply the methodology used by \newcite{shoemark2017} in the context of the 2014 Scottish independence referendum to a dataset of tweets related to the Catalonian referendum.
We use the phenomenon of \textit{code-switching} between Catalan and Spanish to pursue the following research questions in order to understand the choice of language in the context of the referendum:
\begin{enumerate}
\setlength\itemsep{0pt}
\item Is a speaker's stance on independence strongly associated with the rate at which they use Catalan?
\item Does Catalan usage vary depending on whether the discussion topic is related to the referendum, and on the intended audience?
\end{enumerate}
For the first question, our findings are similar to those in the Scottish case: pro-independence tweets are more likely to be written in Catalan than anti-independence tweets, and pro-independence Twitter users are more likely to use Catalan than anti-independence Twitter users (Section \ref{sec:exp1}).
With respect to the second question, we find that Twitter users are more likely to use Catalan in referendum-related tweets, and that they are more likely to use Catalan in tweets with a broader audience (Section \ref{sec:exp2}).\footnote{Code for collecting data and rerunning the experiments is available at \url{https://github.com/ianbstewart/catalan}.}

\section{Related work}
Code-switching, the alternation between languages within conversation~\cite{poplack1980}, has been shown to be the product of grammatical factors, such as syntax~\cite{pfaff1979}, and social factors, such as intended audience~\cite{gumperz1977}.
While many studies have examined code-switching in the spoken context~\cite{auer2013}, social media platforms such as Twitter provide an opportunity to study code-switching in online discussions~\cite{androutsopoulos2015}.
In the online context, choice of language may reflect the writer's intended audience~\cite{kim2014} or identity~\cite{christiansen2015,lavendar2017}, and the explicit social signals in online discussions such as @-replies can be leveraged to test claims about code-switching at a large scale~\cite{nguyen2015}.

A relatively unexplored area of code-switching behavior is politically-motivated code-switching, which we assume has a different set of constraints compared to everyday code-switching.
With respect to political separatism, \newcite{shoemark2017} studied the use of Scots, a language local to Scotland, in the context of the 2014 Scotland independence referendum.
They found that Twitter users who openly supported Scottish independence were more likely to incorporate words from Scots in their tweets.
They also found that Twitter users who tweeted about the referendum were less likely to use Scots in referendum-related tweets than in non-referendum tweets.

This study considers the similar scenario which took place in 2017 vis-\`{a}-vis the semi-autonomous region of Catalonia.
Our main methodological divergence from \newcite{shoemark2017} relates to the linguistic phenomenon at hand:
while Scots is mainly manifested as interleaving individual words within English text (code-mixing), Catalan is a distinct language which, when used, usually replaces Spanish altogether for the entire tweet (code-switching).

%% file: t01_hashtag_tables.tex
\begin{table*}[t!]
  \small
  \centering
  \begin{tabular}{l p{12cm}} \toprule
    Neutral &  \#1O (748), \#1Oct (1351), \#1Oct2017 (171), \#1Oct2017votarem (28), \#CatalanRef2017 (46), \#CatalanReferendum (3244), \#CatalanReferendum2017 (72), \#JoVoto (54), \#Ref1oct (90), \#Referéndum (640), \#Referendum1deoctubre (146), \#ReferendumCAT (457), \#ReferendumCatalan (298), \#Votarem (954) \\
    Pro-independence &  \#1ONoTincPor (18), \#1octL6 (184), \#CataloniaIsNotSpain (10), \#CATvotaSí (3), \#CataluñaLibre (27), \#FreePiolin (293), \#Freedom4Catalonia (2), \#IndependenciaCataluña (9), \#LetCatalansVote (3), \#Marxem (102), \#RepúblicaCatalana (212), \#Spainispain (8), \#SpanishDictatorship (9), \#SpanishRepression (3), \#TotsSomCatalunya (261) \\
    Anti-independence & \#CataluñaEsEspaña (69), \#DontDUIt (12), \#EspanaNoSeRompe (29), \#EspañaUnida (4), \#OrgullososDeSerEspañoles (55), \#PorLaUnidadDeEspaña (2), \#ProuPuigdemont (187) \\ \bottomrule
    \end{tabular}
  \caption{Hashtags related to the Catalonian referendum, their stances (neutral/pro/anti) and their frequencies in the CT dataset.}
  \label{tab:ref_hashtags}
\end{table*}

%% file: 02_setup.tex
\section{Data}

The initial set of tweets for this study, $\mathcal{T}$, was drawn from a 1\% Twitter sample mined between January 1 and October 31, 2017, covering nearly a year of activity before the referendum, as well as its immediate aftermath.\footnote{A preliminary check of our data revealed that the earliest referendum discussions began in January, 2017.}

The first step in building this dataset was to manually develop a seed set of hashtags related to the referendum.
Through browsing referendum content on Twitter, the following seed hashtags were selected: \textit{\#CataluñaLibre, \#IndependenciaCataluña, \#CataluñaEsEspaña, \#EspañaUnida}, and \textit{\#CatalanReferendum}.
All tweets containing at least one of these hashtags were extracted from $\mathcal{T}$, and the top 1,000 hashtags appearing in the resulting dataset were manually inspected for relevance to the referendum.
From these co-occurring hashtags, we selected a set of 46 hashtags and divided it into pro-independence, anti-independence, and neutral hashtags, based on translations of associated tweet content.\footnote{Authors have a reading knowledge of Spanish. For edge cases we consulted news articles relating to the hashtag.}
After including ASCII-equivalent variants of special characters, as well as lowercased variants, our final hashtag set comprises 111 unique strings.

Next, all tweets containing any referendum hashtag were extracted from $\mathcal{T}$, yielding 190,061 tweets.
After removing retweets and tweets from users whose tweets frequently contained URLs (i.e., likely bots), our final ``Catalonian Independence Tweets'' (CT) dataset is made up of 11,670 tweets from 10,498 users (cf. the Scottish referendum set IT with 59,664 tweets and 18,589 users in~\newcite{shoemark2017}).
36 referendum-related hashtags appear in the filtered dataset. They are shown with their frequencies (including variants) in~\autoref{tab:ref_hashtags} (cf. the 47 hashtags and similar frequency distribution in Table 1 of~\newcite{shoemark2017}).

To address the control condition, all authors of tweets in the CT dataset were collected to form a set $\mathcal{U}$, and all other tweets in $\mathcal{T}$ written by these users were extracted into a control dataset (XT) of 45,222 tweets (cf. the 693,815 control tweets in Table 6 of~\newcite{shoemark2017}).

The CT dataset is very balanced with respect to the number of tweets per user: only four users contribute over ten tweets (max = 14) and only 16 have more than five.
The XT dataset also has only a few ``power'' users, such that nine users have over 1,000 tweets (max = 3,581) and a total of 173 have over 100 tweets.
Since the results are macro-averaged over all users, these few power users should not significantly distort the findings.

\paragraph{Language Identification.}
\label{ssec:langid}

This study compares variation between two distinct languages, Catalan and Spanish.
We used the \textit{langid} language classification package~\cite{lui2012}, based on character n-gram frequencies,
to identify the language of all tweets in CT and XT.
Tweets that were not classified as either Spanish or Catalan with at least 90\% confidence were discarded.
This threshold was chosen by manual inspection of the \textit{langid} output.
In the referendum dataset CT (control set XT), \emph{langid} confidently labeled 4,014 (56,892) tweets as Spanish and 2,366 (10,178) as Catalan.
To address the possibility of code-mixing within tweets, the first two authors manually annotated a sample of 100 tweets, of which half were confidently labeled as Spanish, and the other half as Catalan.
They found only two examples of potential code-mixing, both of Catalan words in Spanish text.

%% file: 03_exp1.tex
\section{Catalan Usage and Political Stance}
\label{sec:exp1}

The first research question concerns political stance: do pro-independence users tweet in Catalan at a higher rate than anti-independence users?

We analyze the relationship between language use and stance on independence under two conditions, comparing the use of Catalan among pro-independence users vs. anti-independence users in
(1) opinionated referendum-related tweets (tweets with Pro/Anti hashtags); and
(2) all tweets.
These conditions address the possibilities that the language distinction is relevant for pro/anti-independence Twitter users in political discourse and outside of political discourse, respectively.

\paragraph{Method.}
The first step is to divide the Twitter users in $\mathcal{U}$ into pro-independence (\textit{PRO}) and anti-independence (\textit{ANTI}) groups.
First, the proportion of tweets from each user that include a pro-independence hashtag is computed as $\frac{\npro}{\npro + \nanti}$, where $\npro$ ($\nanti$) is the count of tweets from user $u$ that contain a pro- (anti-) independence hashtag.
The \textit{PRO} user set ($\pro{\mathcal{U}}$) includes all users whose pro-independence proportion was above or equal to 75\%, and the \textit{ANTI} user set ($\anti{\mathcal{U}}$) includes all users whose pro-independence proportion was below or equal to 25\%.
The counts of users and tweets identified as either Spanish or Catalan are presented in~\autoref{tab:exp_1_stats}.

\begin{table}
  \small
  \centering
  \begin{tabular}{l r r r r} \toprule
    ~ & \multicolumn{2}{c}{Tweets with} & \multicolumn{2}{c}{All tweets}\\
    ~ & \multicolumn{2}{c}{Pro/Anti hashtags} & ~ & ~ \\ \midrule
    Group & \emph{PRO} & \emph{ANTI} & \emph{PRO} & \emph{ANTI} \\
    \# Users & 713 & 242 & 1,011 & 312 \\
    \# Tweets & 858 & 288 & 44,229 & 22,841 \\ \bottomrule
  \end{tabular}
  \caption{Tweet and user counts for the stance study.}
  \label{tab:exp_1_stats}
\end{table}

\begin{table}
  \small
  \centering
  \begin{tabular}{l r r} \toprule
    ~ & Tweets with & All tweets \\
     & Pro/Anti hashtags & \\ \midrule
    $\pro{\hat{p}}$ & 0.3136 & 0.2772 \\
    $\anti{\hat{p}}$ & 0.0613 & 0.0586 \\
    $d$ & 0.2523 & 0.2186 \\
    $p$-value & $< 10^{-5}$ & $< 10^{-5}$ \\ \bottomrule
  \end{tabular}
\caption{Results of the stance study. $d = \pro{\hat{p}} - \anti{\hat{p}}$.}
\label{tab:exp_1_results}
\end{table}

To measure Catalan usage,
let $\nca$ and $\nes$ denote the counts of Catalan and Spanish tweets user $u$ posted, respectively.
We quantify Catalan usage using the proportion $\hat{p}^{(u)} = \frac{\nca}{\nca + \nes}$, computing the macro-average over each group $\mathcal{U}_{G}$'s members to produce $\hat{p}_{G} = \frac{1}{|\mathcal{U}_{G}|} \sum_{u \in \mathcal{U}_{G}} \hat{p}^{(u)}$.
The test statistic is then the difference in Catalan usage between the pro- and anti-independence groups, 
$d = \pro{\hat{p}} - \anti{\hat{p}}$.

To determine significance, the users are randomly shuffled between the two groups to recompute $d$ over 100,000 iterations.
The $p$-value is the proportion of permutations in which the randomized test statistic was greater than or equal to the original test statistic from the unpermuted data.

\paragraph{Results.}
Catalan is used more often among the pro-independence users compared to the anti-independence users, across both the hashtag-only and all-tweet conditions.
\autoref{tab:exp_1_results} shows that the proportion of tweets in Catalan for pro-independence users ($\pro{\hat{p}}$) is significantly higher than the proportion for anti-independence users ($\anti{\hat{p}}$).
This is consistent with \newcite{shoemark2017}, who found more Scots usage among pro-independence users ($d=0.00555$ for pro/anti tweets, $d=0.00709$ for all tweets).
The relative differences between the groups are large: in the all-tweet condition, $\pro{\hat{p}}$ is five times greater than $\anti{\hat{p}}$, whereas \citeauthor{shoemark2017} found a twofold difference ($\pro{\hat{p}}=0.01443$ versus $\anti{\hat{p}}=0.00734$ for all-tweet condition).
All raw proportions are two orders of magnitude greater than those in the Scottish study, a result of the denser language variable used in this study (full-tweet code-switching vs. intermittent code-mixing).

%% file: 04_exp2.tex
\section{Catalan Usage, Topic, and Audience}
\label{sec:exp2}

One way to explain the variability in Catalan usage is through \emph{topic-induced variation}, which proposes that people adapt their language style in response to a shift in topic~\cite{rickford1994}.
This leads to our second research question: is Catalan more likely to be used in discussions of the referendum than in other topics?
This analysis is conducted under three conditions.
The first two conditions compare Catalan usage in referendum-hashtag tweets (pro, anti, and neutral) against (1) all tweets; and (2) tweets that contain a non-referendum hashtag.
This second condition is meant to control for the general role of hashtags in reaching a wider audience~\cite{pavalanathan2015}, and its results motivate the third analysis, comparing (3) @-reply tweets with hashtag tweets.

\subsection{Referendum Hashtags}

\paragraph{Method.} We extract all users in $\mathcal{U}$ who have posted at least one referendum-related tweet and at least one tweet unrelated to the referendum into a new set, $\mathcal{U}_R$.
Tweet and user counts for all conditions are provided in \autoref{tab:exp_2_stats}.
The small numbers are a result of the condition requirement and the language constraint (tweets must be identified as Spanish or Catalan with 90\% confidence).
For a user $u$, we denote the proportion of $u$'s referendum-related tweets written in Catalan by $\phu_{C}$, and the proportion of $u$'s control tweets written in Catalan by $\phu_{X}$.
We are interested in the difference between these two proportions $d^{(u)}=\phu_{C} - \phu_{X}$ and its average across all users $\bar{d}_{\mathcal{U}_R}=\frac{1}{|\mathcal{U}_R|} \sum_{u \in \mathcal{U}_R} d^{(u)}$.
Under the null hypothesis that Catalan usage is unrelated to topic, $\bar{d}_{\mathcal{U}_R}$ would be equal to $0$, which we test for significance using a one-sample t-test.

\paragraph{Results.}
Our results, presented in the middle columns of \autoref{tab:exp_2_results}, show that users tweet in Catalan at a significantly higher rate in referendum tweets than in all control tweets (first results column), but no significant difference was observed in the control condition where tweets include at least one hashtag (second results column).
The lack of a significant difference between referendum-related hashtags and other hashtags suggests that the topic being discussed is not as central in choosing one's language, compared with the audience being targeted.

\begin{table}
  \small
  \centering
  \begin{tabular}{l r r r} \toprule
    Treatment set & Ref. hash & Ref. hash & Replies \\
     Control set & All tweets & All hash & All hash
    \\ \midrule
    Users & 772 & 548 & 654 \\ 
    Treatment tweets & 887 & 656 & 6225 \\
    Control tweets & 31,151 & 13,954 & 10,319 \\ \bottomrule
  \end{tabular}
   \caption{Tweet and user counts for each condition in the topic/audience study. `hash' stands for `tweets with hashtags'.}
  \label{tab:exp_2_stats}
\end{table}

\begin{table}
  \small
  \centering
  \begin{tabular}{l r r r} \toprule
    Treatment set & Ref. hash & Ref. hash & Replies \\
     Control set & All tweets & All hash & All hash 
     \\ \midrule
    $\bar{d}_{\mathcal{U}_R}$ & $0.033$ & $0.018$ & $-0.031$ \\ 
    Standard error & $0.011$ & $0.011$ & $0.011$ \\
    $t$-statistic & $3.02$ & $1.59$ & $-2.79$ \\
    $p$-value & $0.002$ & $0.111$ & $0.005$ \\ \bottomrule
  \end{tabular}
  \caption{Results of the topic/audience study. $\bar{d}_{\mathcal{U}_{R}}$ is the difference in rate of Catalan use between treatment settings and control settings, averaged across users.}
  \label{tab:exp_2_results}
\end{table}

Our second result is the opposite of the prior finding that there were significantly \textit{fewer} Scots words in referendum-related tweets than in control tweets (cf. Table 7 in \newcite{shoemark2017}; $\bar{d}_{u}=-0.0015$ for all controls).
This suggests that Catalan may serve a different function than Scots in terms of political identity expression.
Rather than suppressing their use of Catalan in broadcast tweets, users increase their Catalan use, perhaps to signal their Catalonian identity to a broader audience.
This is supported by literature highlighting the integral role Catalan plays in the Catalonian national narrative \cite{crameri2017}, as well as the relatively high proportion of Catalan speakers in Catalonia: 80.4\% of the population has speaking knowledge of Catalan~\cite{catalonianLanguageSurvey}, versus 30\% population of Scotland with speaking knowledge of Scots~\cite{scottishLanguageSurvey}.
There are also systemic differences between the political settings of the two cases:
the Catalonian referendum had much larger support for separation among those who voted (92\% in Catalonia vs. 45\% in Scotland)~\cite{fotheringham2017,jeavens2014}.
These factors suggest a different public perception of national identity in the two regions within the context of the referenda, resulting in different motivations behind language choice.

\subsection{Reply Tweets}
Earlier work has highlighted the role of hashtags and @-replies as affordances for selecting large and small audiences, and their interaction with the use of non-standard vocabulary~\cite{pavalanathan2015}. To test the role of audience size in Catalan use, we compare the proportion of Catalan in @-reply tweets against hashtag tweets.

\paragraph{Method.} In this analysis, we take the treatment set to be all tweets made by users in $\mathcal{U}_R$ which contain an @-reply but not a hashtag (narrow audience), and control against all tweets which contain a hashtag but not an @-reply (wide audience).

\paragraph{Results.} The results in the rightmost column of \autoref{tab:exp_2_results} demonstrate a significant tendency toward less Catalan use in @-replies than in hashtag tweets.
This trend supports the hypothesis that Catalan is intended for a wider audience.

This effect may also be explained by a subset of reply tweets in political discourse being targeted at national figures, possibly seeking to direct the message at the target's followers rather than to engage in discussion with the target.
For example, one of the reply-tweets addresses a Spanish politician (``user1'') in a conversation about a recent court case: \tweetquote{@user1 @user2 What justice are you talking about? What can a JUDGE like this impart?}\footnote{@user1 @user2 De que justícia hablas? De la que pueda impartir un JUEZ como este?}.
The same writer uses Catalan in a more broadcast-oriented message: \tweetquote{Enough [being] dumb! We'll get to work and do not divert us from our way. First independence, then what is needed! My part; \#CatalonianRepublic}\footnote{Prou rucades! Anem per feina i no ens desviem del camí. El primer la independència, després el que calgui! El meu parti; \#republicacatalana}. 
This provides a new perspective on the earlier finding by \newcite{pavalanathan2015}: by replying to tweets from well-known individuals, it may be possible to reach a large audience, similar to the use of popular hashtags.

%% file: 05_conclusion.tex
\section{Conclusion}
\label{sec:disc}
This study demonstrates the association of code-switching with political stance, topic and audience, in the context of a political referendum.
We corroborate prior work by showing that the use of a minority language is associated with pro-independence political sentiment, and we also provide a result in contrast to prior work, that the use of a minority language is associated with a broader intended audience.
This study extends the setting of code-switching from everyday conversation into specifically political conversation, which is subject to different expectations and constraints.

This study does not use geographic signals, because the sparsity of geotagged tweets prevented us from restricting the scope to data generated in Catalonia proper.
Another potential limitation is that assumption that political hashtags are robust signals for political stance. Other work has shown that political hashtags can be co-opted by opposing parties~\cite{stewart2017}.

Our findings extend prior work on political use of Scots words on the inter-speaker level and Scots-English code-mixing on the intra-speaker level to examining language choice and code-switching, respectively.
Further work is required to reconcile our results with prior work on topic differences and audience size~\cite{pavalanathan2015}.
Future work may also compare the Catalonian situation with multilingual societies in which a minority language is discouraged~\cite{karrebaek2013}, or in which the languages are more equally distributed~\cite{blommaert2011}.

%% file: 06_acknowledgments.tex
\section*{Acknowledgments}

We thank Sandeep Soni, Umashanthi Pavalanathan, our anonymous reviewers, and members of Georgia Tech's Computational Social Science class for their feedback.
This research was supported by NSF award IIS-1452443 and NIH award R01-GM112697-03.